\documentclass[conference]{IEEEtran}
\IEEEoverridecommandlockouts
\usepackage{subfigure}

\usepackage{cite}
\usepackage{tabularx}
\usepackage{amsmath,amssymb,amsfonts}
\usepackage{algorithmic}
\usepackage{graphicx}
\usepackage{textcomp}
\usepackage{booktabs}
\usepackage{xcolor}
\usepackage[capitalize]{cleveref}
\usepackage{color}
\definecolor{plum}{RGB}{142,69,133}

\def\BibTeX{{\rm B\kern-.05em{\sc i\kern-.025em b}\kern-.08em
    T\kern-.1667em\lower.7ex\hbox{E}\kern-.125emX}}

\makeatletter
\newcommand{\linebreakand}{%
  \end{@IEEEauthorhalign}
  \hfill\mbox{}\par
\mbox{}\hfill\begin{@IEEEauthorhalign}
}
\makeatother

\begin{document}

\title{Char-SAM: Turning Segment Anything Model into Scene Text Segmentation Annotator with Character-level Visual Prompts}

\author{\IEEEauthorblockN{ Enze Xie}
\IEEEauthorblockA{\textit{Institute of Information Engineering} \\
\textit{Chinese Academy of Sciences}\\
\textit{School of Cyber Security,
University} \\ 
\textit{of Chinese Academy of Sciences}\\
Beijing, China \\
xieenze@iie.ac.cn}
\and
\IEEEauthorblockN{Jiahao Lyu}
\IEEEauthorblockA{\textit{Institute of Information Engineering} \\
\textit{Chinese Academy of Sciences}\\
\textit{School of Cyber Security,
University} \\ 
\textit{of Chinese Academy of Sciences}\\
Beijing, China \\
lvjiahao@iie.ac.cn}
\and
\IEEEauthorblockN{Daiqing Wu}
\IEEEauthorblockA{\textit{Institute of Information Engineering} \\
\textit{Chinese Academy of Sciences}\\
\textit{School of Cyber Security,
University} \\ 
\textit{of Chinese Academy of Sciences}\\
Beijing, China \\
wudaiqing@iie.ac.cn}
\linebreakand
\IEEEauthorblockN{Huawen Shen}
\IEEEauthorblockA{\textit{Institute of Information Engineering} \\
\textit{Chinese Academy of Sciences}\\
\textit{School of Cyber Security,
University} \\ 
\textit{of Chinese Academy of Sciences}\\
Beijing, China \\
shenhuawen@iie.ac.cn}
\and  
\IEEEauthorblockN{Yu Zhou$^{\ast}$ \thanks{* Corresponding author}}
\IEEEauthorblockA{
\textit{VCIP \& TMCC \& DISSec} \\
\textit{College of Computer Science}\\ 
\textit{
Nankai University} \\
Tianjin, China \\
yzhou@nankai.edu.cn
}
}
\maketitle

\begin{abstract}
The recent emergence of the Segment Anything Model (SAM) enables various domain-specific segmentation tasks to be tackled cost-effectively by using bounding boxes as prompts. However, in scene text segmentation, SAM can not achieve desirable performance. The word-level bounding box as prompts is too coarse for characters, while the character-level bounding box as prompts suffers from over-segmentation and under-segmentation issues. 
In this paper, we propose an automatic annotation pipeline named Char-SAM, that turns \textit{SAM} into a low-cost segmentation annotator with a \textit{Char}acter-level visual prompt. Specifically, leveraging some existing text detection datasets with word-level bounding box annotations, we first generate finer-grained character-level bounding box prompts using the Character Bounding-box Refinement (\textbf{CBR}) module. Next, we employ glyph information corresponding to text character categories as a new prompt in the Character Glyph Refinement (CGR) module to guide SAM in producing more accurate segmentation masks, addressing issues of over-segmentation and under-segmentation.  
These modules fully utilize the bbox-to-mask capability of SAM to generate high-quality text segmentation annotations automatically.
Extensive experiments on TextSeg validate the effectiveness of Char-SAM. Its training-free nature also enables the generation of high-quality scene text segmentation datasets from real-world datasets like COCO-Text and MLT17.



\end{abstract}


\section{Introduction}

          
          
          
          

\begin{figure}[t]
    \centering
    \includegraphics[width=\linewidth]{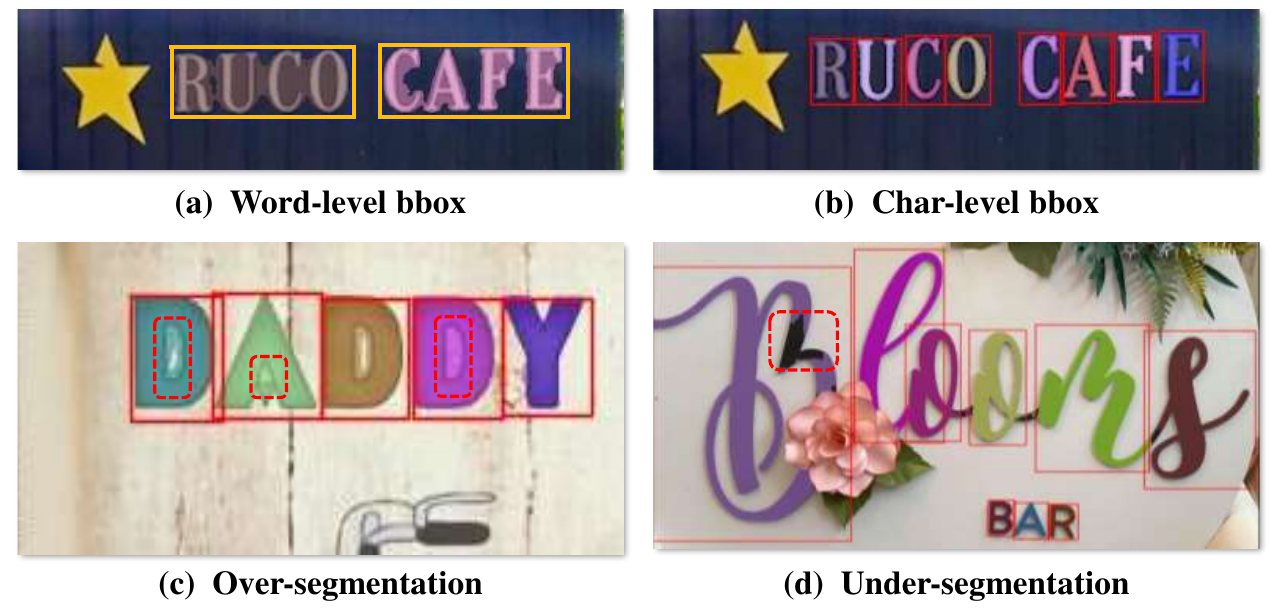}
    \caption{Failure cases of SAM on TextSeg \cite{xu2021rethinking}. The yellow and red rectangles in (a) and (b) represent the word-level and character-level bounding box prompts provided to SAM respectively. Red rectangular with dashed lines in (c) and (d) highlight the failure segmentations. Zoom in and out for a better view.}
    \label{badcase}
    \vspace{-15pt}
\end{figure}

\begin{figure*}[t]
    \centering
    \includegraphics[width=1\linewidth]{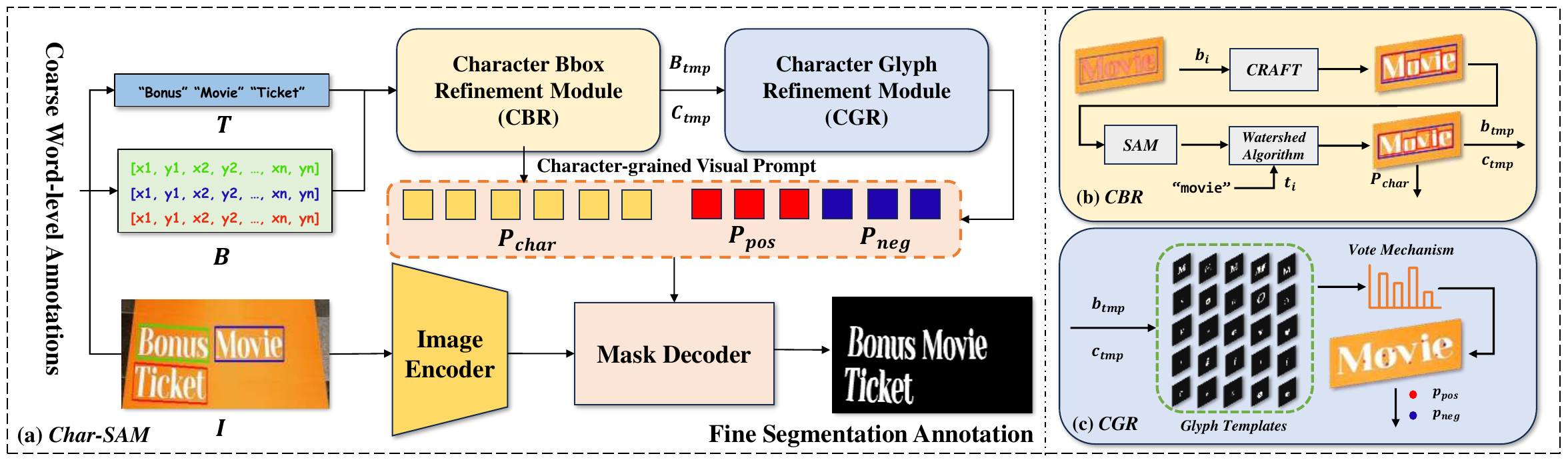}
    \caption{ The overall framework of our \textbf{Char-SAM}, which mainly consists of the Character Bbox
Refinement (\textbf{CBR}) module, the Character Glyph
Refinement (\textbf{CGR}) module and SAM architecture.}
    \label{framework}
    \vspace{-15pt}
\end{figure*}

In the field of scene text research, tasks such as detection\cite{shu2023perceiving, chen2021self}, recognition\cite{qiao2020seed, qiao2021pimnet}, spotting\cite{lyu2024arbitrary, wang2022tpsnet}, processing\cite{zeng2024textctrl, li2024first}, VQA\cite{zhang2024track, zeng2023beyond}, and understanding\cite{shen2024ldp, shen2023divide} are included. In addition, scene text segmentation has also gradually gained attention. Compared with word-level bounding boxes, scene text segmentation\cite{tft,xu2021rethinking,bonechi2019coco_ts,zhao2017pyramid,bonechi2020weak,ye2024hi, Ren_2022_ACCV} generates fine-grained pixel-level masks that are more beneficial for downstream tasks, including scene text erasure\cite{texteraser} and scene text editing\cite{textedit}.
However, the expensive cost of pixel-level annotations results in an extreme lack of high-quality datasets for scene text segmentation. 

Due to the labor-intensive annotation, pseudo-annotated datasets are often utilized for training scene text segmentation models. For instance, COCO\_TS\cite{bonechi2019coco_ts} and MLT\_S\cite{bonechi2020weak} are derived from scene text detection datasets COCO-Text\cite{veit2016coco} and MLT17\cite{nayef2017icdar2017} respectively. However, the previous pseudo-annotation process predominantly employs semi-supervised algorithms and suffers from fragmented text masks and misclassification of non-text pixels. These issues underscore the necessity of developing a scene text segmentation annotation pipeline that wields both high performance and low expense.

As a powerful prompt-based vision foundation model for image segmentation, Segment Anything Model (SAM) \cite{SAM} can accurately segment objects of interest in an image based on diverse user-input prompts, including point, bounding box, text, and mask. 
By learning from a large volume of high-quality masks in the SA-1B dataset \cite{SAM}, SAM obtains superior generalizability across various domains.
Therefore, many recent works have migrated the ability of SAM to domain-specific segmentation tasks such as medical segmentation \cite{ma2024segment, wu2023medical}, anomaly segmentation\cite{li2024clipsam} and remote sensing segmentation \cite{chen2024rsprompter}, achieving remarkable performance. 

However, applying SAM directly to scene text segmentation reveals two main issues in TextSeg\cite{xu2021rethinking} images: 1) \textbf{Insufficient Prompt Granularity}. 
Solely providing word-level bounding boxes is insufficient for SAM to segment individual characters accurately. As shown in \Cref{badcase}(a), SAM treats characters as a united entity rather than separate ones. Therefore, character-level bounding box prompts are required, as illustrated in \Cref{badcase}(b).
2) \textbf{Over- and Under-segmentation.} Even with a character-level bounding box, SAM struggles to distinguish the hole area  (background area surrounded by foreground area) from the text area, resulting in over-segmentation in the hole area, as shown in \Cref{badcase}(c). 
This issue arises because the masks in SA-1B \cite{SAM} with holes only account for a small part, while many text characters contain hole areas such as 'D', 'A', \textit{etc}. 
Additionally, \Cref{badcase}(d) shows that the segmentation output of SAM could obtain an incomplete mask, especially when text characters are of large scale.

Based on the above observations and analysis, we propose a framework named \textbf{Char-SAM}, which derives from the \textbf{Char}acter-level visual prompt \textbf{S}egment \textbf{A}nything \textbf{M}odel. The overall pipeline consists of the Character Bounding-box Refinement (\textbf{CBR}) module and Character Glyph Refinement (\textbf{CGR}) module. The \textbf{CBR} module refines word-level annotations to more fine-grained character-level annotations, and the \textbf{CGR} module employs glyph information of text characters to tackle over-segmentation and under-segmentation. Noteworthy, the whole process is training-free.
Extensive experiments show that the proposed framework achieves competitive performance and outperforms other methods for generating text segmentation annotations, demonstrating the generality of our proposed approach.
The main contributions of this paper can be summarized as follows:

\begin{enumerate}
    \item Leveraging the powerful segmentation ability of SAM, we propose a novel automatic annotation pipeline \textbf{Char-SAM} for scene text segmentation via character-level visual prompts, without any external training process. 
    \item To adapt the scene text segmentation task, \textbf{CBR} and \textbf{CGR} modules are designed to obtain fine-grained prompts and solve over-/ under-segmentation issues respectively.
    \item 
    Based on real-world datasets COCO-Text and MLT17, we generate character-level text segmentation datasets COCO\_TS\_refined and MLT\_S\_refined, which benefits the performance for scene text segmentation.
\end{enumerate}

\section{Method}
\subsection{Framework}
Char-SAM aims to refine word-level annotations corresponding to scene images into pixel-level annotations.
As illustrated in \Cref{framework}, the input composes of an input image $I\in \mathbb{R}^{H\times W\times 3}$, word bounding boxes $B = \{b_i\}_{i=1}^n$ and text transcriptions $T= \{t_i\}_{i=1}^n$, where $b_i$ is the $i$-th input word bounding box and $t_i$ indicates the corresponding transcription.
The \textbf{CBR} module aims to obtain the precise character-level bounding boxes $B_{tmp}$ and its corresponding character categories $C_{tmp}$. The refined character bounding boxes are regarded as the visual prompts $P_{char}$. After feeding each input character-level bounding box $b_{tmp}\in B_{tmp}$ and character category $c_{tmp}\in C_{tmp}$, the \textbf{CGR} module uses the glyph information corresponding to the category to generate positive point prompts $P_{pos}$ and negative point prompts $P_{neg}$. Finally, the input image and the fine-grained prompts including $P_{char}, P_{pos}$ and $P_{neg}$ are fed together into SAM to generate accurate segmentation annotations.

\subsection{Character Bounding-box Refinement Module}
Scene text detection datasets\cite{ch2017total, singh2021textocr,veit2016coco,nayef2017icdar2017,chng2019icdar2019} usually provide word-level bounding boxes and transcriptions. However, preliminary results in \Cref{badcase} show that word-level prompts are insufficient for SAM to achieve accurate segmentation. Therefore, the \textbf{CBR} module is proposed to refine word-level annotations to character-level. 

Specifically, we firstly use the character-awareness text detector CRAFT\cite{craft} trained on SynthText\cite{synthtext} dataset for preliminary detection. However, due to the domain gap between synthetic and real-world images, some adjacent character boxes cannot be completely separated, as exemplified by "Movie" in \Cref{framework}(b). This will fail to uniquely assign a character category to a bounding box, and the glyph information corresponding to the character category cannot be used.
To address this issue, we further determine whether the current character bounding box is detached based on the number of characters in the input transcription $t_i$.
Subsequently, we use the bounding boxes to crop the input image and feed the cropped regions into the text recognizer to establish a one-to-one correspondence between character-level boxes and character categories.
For sequences length greater than one character, such as the ``vi" of ``movie" in \Cref{framework}(b), we feed the bounding box and the image into SAM and use the watershed algorithm to partition the logical map of the corresponding area of the bounding box into finer segments.
Finally, the character-level bounding box is calculated through the connected domain, and an elaborated bipartite matching between character boxes and categories is implemented.
          
          
        

\subsection{Character Glyph Refinement Module}
Through the \textbf{CBR} module, we obtain both the bounding box and category for each character, whereas the mask obtained by feeding the character-level box as a prompt into SAM still has problems such as over-segmentation and under-segmentation illustrated as \Cref{badcase}. We observe that the English character, regardless of font or text direction, follows the fixed glyph structure determined by its category. Therefore, by utilizing known text categories across various English fonts, we can generate corresponding glyph templates and refine segmentation results using glyph information. 

In detail, based on the collected text fonts, we can generate glyph templates for character categories with the $ImageDraw$ and $ImageFont$ library functions. We then categorize the pixels of these glyph templates into foreground and hole pixels. Each glyph template casts votes for each category, with pixels receiving votes based on their frequency. It is worth noting that only pixels with a vote rate exceeding a threshold are retained. Hole pixels become false point prompts $P_{neg}$, while the foreground pixels become positive point prompts $P_{seg}$. Finally, these visual prompts are fed into SAM to produce the final refined mask. 

\section{Experiments}
\label{sec:experiment}
\subsection{Datasets}
\textbf{TextSeg} consists of 4024 high-quality images of scene and artistic text, with annotations including word-level and character-level bounding boxes and character-level masks.

\textbf{COCO\_TS} contains 14690 low-quality pixel-level annotations for images in COCO-Text dataset. The annotations are obtained from the available bounding–boxes of the COCO–Text dataset exploiting a weakly supervised algorithm.

\textbf{MLT\_S} contains 6896 low-quality pixel-level annotations of MLT images. Similarly, the annotations are obtained by a semi-supervised algorithm using bounding boxes.




\begin{figure*}[t]
    \centering
    \includegraphics[width=1\linewidth]{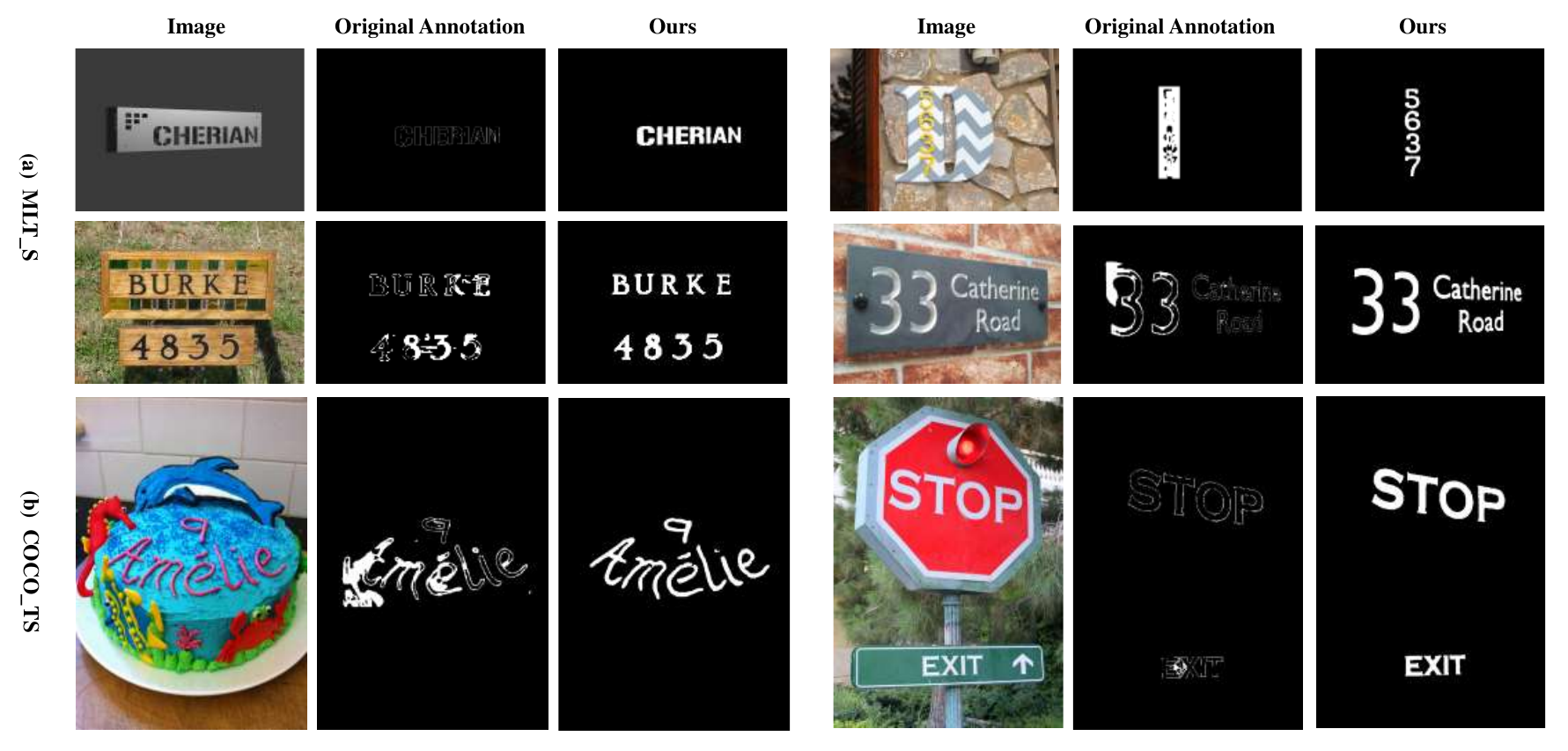}
    \caption{ Comparison of annotation quality of different scene text segmentation datasets.}
    \label{visualization}
    \vspace{-15pt}
\end{figure*}

\subsection{Implementation Details}
We choose SAM-B as our baseline, with the lightweight ViT-B\cite{dosovitskiy2020image} as the image encoder. The threshold in the \textbf{CGR} module is set to 0.6 by default. The number of text font files collected is 80.
 We conduct ablation experiments on the TextSeg dataset.

\subsection{ Comparison with SOTA Approaches}
To demonstrate the accuracy of our method on scene text segmentation, we compare the performance of Char-SAM with other supervised scene text segmentation SOTA models on TextSeg as shown in \Cref{sota}. While maintaining single-scale input resolution, Char-SAM still has competitive performances compared to supervised models with the multi-scale evaluation strategy. For the F-score, it is only about 1\% behind the SOTA model TFT\cite{tft}. It proves our pipeline is solid on automatic annotation in scene text segmentation.

\begin{table}[t]
\centering
\caption{Comparison of State-of-the-art Performances on TextSeg dataset.${\dagger}$ means the multi-scale evaluation strategy is employed to improve the performance of the inference.}
\begin{tabular}{lcc}
\toprule
Methods                & fgIoU(\%) & F-score(\%) \\ \hline
\multicolumn{3}{c}{Supervised Performance}
\\ \hline
PSPNet \cite{bonechi2019coco_ts,bonechi2020weak}                & -         & 74.00       \\
SMANet  \cite{zhao2017pyramid}               & -         & 77.00       \\
TextRNet + DeepLabV3+$^{\dagger}$\cite{xu2021rethinking,chen2017deeplab}  & 86.06     & 92.10       \\
TextRNet + HRNetV2-W58$^{\dagger}$ \cite{xu2021rethinking,hrnetv2} & 86.84     & 92.40       \\
TFT$^{\dagger}$  \cite{tft}                  & 87.11     & 93.10       \\ \hline
\multicolumn{3}{c}{Zero-shot Performance}
\\ \hline
Char-SAM (Ours)               & 84.80     & 92.15       \\ \bottomrule
\end{tabular}

\label{sota}
\end{table}

\begin{table}[t]
\centering
\caption{Ablation study of proposed modules on TextSeg dataset. ``TN" indicates the template Number.}
\label{module}
\begin{tabular}{lcccc}
\toprule
      Settings &
      CBR & CGR &  fgIoU(\%) &  F-score(\%) \\ \midrule 
Baseline     & - & - & 78.30                        & 83.52                          \\
\#1        & \checkmark & - & 81.85                        & 87.81                          \\
\#2 (TN=40) & \checkmark & \checkmark & 84.75                        & 92.01                          \\
\#3 (TN=80) & \checkmark & \checkmark & 84.80                        & 92.15 \\
\bottomrule
\end{tabular}
\vspace{-10pt}
\end{table}
\subsection{Ablation Study}
\subsubsection{Key modules}
To evaluate the effectiveness of \textbf{CBR} and \textbf{CGR}, we conduct ablation studies on these modules, as shown in \Cref{module}. According to the baseline and \#1, the \textbf{CBR} module can increase fgIoU and F-score by more than 3\%. Illustrated by \#1 and \#2, the \textbf{CGR} module provides more fine-grained prompts based on glyph information, improving fgIoU and F-score by about 3\% and 4\% respectively. We also prove that the more templates provided, the more fully the font information can be utilized according to \#2 and \#3.
\subsubsection{Prompt granularity}
We conduct ablation experiments on various prompt types to demonstrate that finer granularity leads to better segmentation performance. As shown in \Cref{prompt}, character-level boxes have significant performance improvement compared to word-level boxes. Moreover, using prompts with positive and negative points can bring further improvement. 
\subsubsection{Dataset quality}
To demonstrate that the quality of the dataset generated by our framework is better than COCO\_TS and MLT\_S generated by semi-supervised methods. We use TextRNet+DeepLabV3+\cite{xu2021rethinking} trained on TextSeg as the baseline. \#1 represents adding a 1000 subset of COCO\_TS and a 1000 subset of MLT\_S to the baseline for training. \#2 indicates replacing annotations of the subsets in \#1 with annotations generated by Char-SAM. As shown in \Cref{trainset}, training with low-quality annotations will significantly reduce model performance by 3.96\% in fgIoU and 2.90\% in F-score, while training with annotations obtained by Char-SAM has a higher performance by 0.65\% in F-score compared with the baseline.



\begin{table}[t]
\setlength{\tabcolsep}{4pt}
\centering\caption{Comparison of SAM's performance under prompts of different granularities on TextSeg dataset. }
\begin{tabular}{lcccc}
\toprule
Prompt Types          & fgIoU(\%) $\uparrow$ &     $\Delta$(\%)  & F-score(\%) $\uparrow$ & $\Delta$(\%)      \\ \midrule
$B$            & 78.30     & -     & 83.52       & -     \\
$P_{char}$           & 82.07     & +3.77 & 88.06       & +4.54 \\
$P_{char}$ + $P_{pos}$       & 83.52     & +5.22 & 91.14       & +7.62 \\
$P_{char}$ + $P_{pos}$ + $P_{neg}$ & 84.68     & +6.38 & 91.91       & +8.39 \\ \bottomrule
\end{tabular}
\label{prompt}
\vspace{-15pt}
\end{table}

\subsection{Visualizations}
We apply our framework on COCO-Text\cite{veit2016coco} and MLT17\cite{nayef2017icdar2017} datasets respectively. The comparison of the obtained pixel-level annotations with COCO\_TS and MLT\_S is shown in \Cref{visualization}. The visualization shows that the masks obtained by our method are more complete and smoother, with fewer misclassifications of foreground and background pixels.

\begin{table}[t]
\centering
\caption{Comparison on TextSeg test set under different settings of training sets.}
\label{trainset}
\begin{tabular}{lcccc}
\toprule
      Training sets   &  fgIoU(\%) & $\Delta$(\%) &  F-score(\%) & $\Delta$(\%) \\ \midrule 
Baseline      & 86.84       &      -           & 92.40           &        -       \\

\#1   &     82.88     &        -3.96       &     89.50   &  -2.90                 \\
\#2   &     86.97     &     +0.13        &  93.05   & +0.65\\
\bottomrule
\end{tabular}
\vspace{-10pt}
\end{table}

\section{Conclusion}
In order to settle the problem of the expensive annotation cost of the scene text segmentation dataset, we propose a framework that can refine word-level annotations into pixel-level annotations. At the same time, considering various issues of SAM on scene text, we proposed \textbf{CBR} and \textbf{CGR} modules to obtain character-level visual prompts. Experiments show that the quality of annotations generated by our Char-SAM is much better than that of semi-supervised methods.

\section*{Acknowledgment}

Supported by the National Natural Science Foundation of China (Grant NO 62376266 and 62406318), and by the Key Research Program of Frontier Sciences, CAS (Grant NO ZDBS-LY-7024).

\section*{}

\bibliographystyle{IEEEtran}
\bibliography{refs}

\begin{thebibliography}{10}
\providecommand{\url}[1]{#1}
\csname url@samestyle\endcsname
\providecommand{\newblock}{\relax}
\providecommand{\bibinfo}[2]{#2}
\providecommand{\BIBentrySTDinterwordspacing}{\spaceskip=0pt\relax}
\providecommand{\BIBentryALTinterwordstretchfactor}{4}
\providecommand{\BIBentryALTinterwordspacing}{\spaceskip=\fontdimen2\font plus
\BIBentryALTinterwordstretchfactor\fontdimen3\font minus \fontdimen4\font\relax}
\providecommand{\BIBforeignlanguage}[2]{{%
\expandafter\ifx\csname l@#1\endcsname\relax
\typeout{** WARNING: IEEEtran.bst: No hyphenation pattern has been}%
\typeout{** loaded for the language `#1'. Using the pattern for}%
\typeout{** the default language instead.}%
\else
\language=\csname l@#1\endcsname
\fi
#2}}
\providecommand{\BIBdecl}{\relax}
\BIBdecl

\bibitem{xu2021rethinking}
X.~Xu, Z.~Zhang, Z.~Wang, B.~Price, Z.~Wang, and H.~Shi, ``Rethinking text segmentation: A novel dataset and a text-specific refinement approach,'' in \emph{Proceedings of the IEEE/CVF Conference on Computer Vision and Pattern Recognition}, 2021, pp. 12\,045--12\,055.

\bibitem{shu2023perceiving}
Y.~Shu, W.~Wang, Y.~Zhou, S.~Liu, A.~Zhang, D.~Yang, and W.~Wang, ``Perceiving ambiguity and semantics without recognition: An efficient and effective ambiguous scene text detector,'' in \emph{Proceedings of the 31st ACM International Conference on Multimedia}, 2023, pp. 1851--1862.

\bibitem{chen2021self}
Y.~Chen, W.~Wang, Y.~Zhou, F.~Yang, D.~Yang, and W.~Wang, ``Self-training for domain adaptive scene text detection,'' in \emph{2020 25th International Conference on Pattern Recognition (ICPR)}.\hskip 1em plus 0.5em minus 0.4em\relax IEEE, 2021, pp. 850--857.

\bibitem{qiao2020seed}
Z.~Qiao, Y.~Zhou, D.~Yang, Y.~Zhou, and W.~Wang, ``{SEED}: Semantics enhanced encoder-decoder framework for scene text recognition,'' in \emph{Proceedings of the IEEE/CVF Conference on Computer Vision and Pattern Recognition}, 2020, pp. 13\,528--13\,537.

\bibitem{qiao2021pimnet}
Z.~Qiao, Y.~Zhou, J.~Wei, W.~Wang, Y.~Zhang, N.~Jiang, H.~Wang, and W.~Wang, ``{PIMNet}: A parallel, iterative and mimicking network for scene text recognition,'' in \emph{Proceedings of the 29th ACM International Conference on Multimedia}, 2021, pp. 2046--2055.

\bibitem{lyu2024arbitrary}
J.~Lyu, W.~Wang, D.~Yang, J.~Zhong, and Y.~Zhou, ``Arbitrary reading order scene text spotter with local semantics guidance,'' in \emph{Proceedings of the AAAI Conference on Artificial Intelligence}, 2025.

\bibitem{wang2022tpsnet}
W.~Wang, Y.~Zhou, J.~Lv, D.~Wu, G.~Zhao, N.~Jiang, and W.~Wang, ``{TPSNet}: Reverse thinking of thin plate splines for arbitrary shape scene text representation,'' in \emph{Proceedings of the 30th ACM International Conference on Multimedia}, 2022, pp. 5014--5025.

\bibitem{zeng2024textctrl}
W.~Zeng, Y.~Shu, Z.~Li, D.~Yang, and Y.~Zhou, ``{TextCtrl}: Diffusion-based scene text editing with prior guidance control,'' \emph{Advances in Neural Information Processing Systems}, 2024.

\bibitem{li2024first}
Z.~Li, Y.~Shu, W.~Zeng, D.~Yang, and Y.~Zhou, ``First creating backgrounds then rendering texts: A new paradigm for visual text blending,'' \emph{ECAI}, 2024.

\bibitem{zhang2024track}
Y.~Zhang, G.~Zeng, H.~Shen, D.~Wu, Y.~Zhou, and C.~Ma, ``Track the answer: Extending {TextVQA} from image to video with spatio-temporal clues,'' in \emph{Proceedings of the AAAI Conference on Artificial Intelligence}, 2025.

\bibitem{zeng2023beyond}
G.~Zeng, Y.~Zhang, Y.~Zhou, X.~Yang, N.~Jiang, G.~Zhao, W.~Wang, and X.-C. Yin, ``Beyond {OCR+VQA}: Towards end-to-end reading and reasoning for robust and accurate {TextVQA},'' \emph{Pattern Recognition}, vol. 138, p. 109337, 2023.

\bibitem{shen2024ldp}
H.~Shen, G.~Li, J.~Zhong, and Y.~Zhou, ``{LDP}: Generalizing to multilingual visual information extraction by language decoupled pretraining,'' in \emph{Proceedings of the AAAI Conference on Artificial Intelligence}, 2025.

\bibitem{shen2023divide}
H.~Shen, X.~Gao, J.~Wei, L.~Qiao, Y.~Zhou, Q.~Li, and Z.~Cheng, ``Divide rows and conquer cells: Towards structure recognition for large tables.'' in \emph{IJCAI}, 2023, pp. 1369--1377.

\bibitem{tft}
H.~Yu, X.~Wang, K.~Niu, B.~Li, and X.~Xue, ``Scene text segmentation with text-focused {T}ransformers,'' in \emph{Proceedings of the 31st ACM International Conference on Multimedia}, 2023, pp. 2898--2907.

\bibitem{bonechi2019coco_ts}
S.~Bonechi, P.~Andreini, M.~Bianchini, and F.~Scarselli, ``{COCO\_TS} dataset: {P}ixel--level annotations based on weak supervision for scene text segmentation,'' in \emph{International Conference on Artificial Neural Networks}.\hskip 1em plus 0.5em minus 0.4em\relax Springer, 2019, pp. 238--250.

\bibitem{zhao2017pyramid}
H.~Zhao, J.~Shi, X.~Qi, X.~Wang, and J.~Jia, ``Pyramid scene parsing network,'' in \emph{Proceedings of the IEEE Conference on Computer Vision and Pattern Recognition}, 2017, pp. 2881--2890.

\bibitem{bonechi2020weak}
S.~Bonechi, M.~Bianchini, F.~Scarselli, and P.~Andreini, ``Weak supervision for generating pixel--level annotations in scene text segmentation,'' \emph{Pattern Recognition Letters}, vol. 138, pp. 1--7, 2020.

\bibitem{ye2024hi}
M.~Ye, J.~Zhang, J.~Liu, C.~Liu, B.~Yin, C.~Liu, B.~Du, and D.~Tao, ``{Hi-SAM}: Marrying segment anything model for hierarchical text segmentation,'' \emph{IEEE Transactions on Pattern Analysis and Machine Intelligence}, pp. 1--16, 2024.

\bibitem{Ren_2022_ACCV}
Y.~Ren, J.~Zhang, B.~Chen, X.~Zhang, and L.~Jin, ``Looking from a higher-level perspective: Attention and recognition enhanced multi-scale scene text segmentation,'' in \emph{Proceedings of the Asian Conference on Computer Vision (ACCV)}, December 2022, pp. 3138--3154.

\bibitem{texteraser}
T.~Nakamura, A.~Zhu, K.~Yanai, and S.~Uchida, ``Scene text eraser,'' in \emph{2017 14th IAPR International Conference on Document Analysis and Recognition (ICDAR)}, vol.~1.\hskip 1em plus 0.5em minus 0.4em\relax IEEE, 2017, pp. 832--837.

\bibitem{textedit}
L.~Wu, C.~Zhang, J.~Liu, J.~Han, J.~Liu, E.~Ding, and X.~Bai, ``Editing text in the wild,'' in \emph{Proceedings of the 27th ACM international conference on multimedia}, 2019, pp. 1500--1508.

\bibitem{veit2016coco}
A.~Veit, T.~Matera, L.~Neumann, J.~Matas, and S.~Belongie, ``{COCO}-{T}ext: Dataset and benchmark for text detection and recognition in natural images,'' \emph{arXiv preprint arXiv:1601.07140}, 2016.

\bibitem{nayef2017icdar2017}
N.~Nayef, F.~Yin, I.~Bizid, H.~Choi, Y.~Feng, D.~Karatzas, Z.~Luo, U.~Pal, C.~Rigaud, J.~Chazalon \emph{et~al.}, ``{ICDAR} 2017 robust reading challenge on multi-lingual scene text detection and script identification-rrc-mlt,'' in \emph{2017 14th IAPR International Conference on Document Analysis and Recognition (ICDAR)}, vol.~1.\hskip 1em plus 0.5em minus 0.4em\relax IEEE, 2017, pp. 1454--1459.

\bibitem{SAM}
A.~Kirillov, E.~Mintun, N.~Ravi, H.~Mao, C.~Rolland, L.~Gustafson, T.~Xiao, S.~Whitehead, A.~C. Berg, W.-Y. Lo \emph{et~al.}, ``Segment anything,'' in \emph{Proceedings of the IEEE/CVF International Conference on Computer Vision}, 2023, pp. 4015--4026.

\bibitem{ma2024segment}
J.~Ma, Y.~He, F.~Li, L.~Han, C.~You, and B.~Wang, ``Segment anything in medical images,'' \emph{Nature Communications}, vol.~15, no.~1, p. 654, 2024.

\bibitem{wu2023medical}
J.~Wu, R.~Fu, H.~Fang, Y.~Liu, Z.~Wang, Y.~Xu, Y.~Jin, and T.~Arbel, ``Medical {SAM} {A}dapter: Adapting segment anything model for medical image segmentation,'' \emph{arXiv preprint arXiv:2304.12620}, 2023.

\bibitem{li2024clipsam}
S.~Li, J.~Cao, P.~Ye, Y.~Ding, C.~Tu, and T.~Chen, ``{ClipSAM}: {CLIP} and {SAM} collaboration for zero-shot anomaly segmentation,'' \emph{arXiv preprint arXiv:2401.12665}, 2024.

\bibitem{chen2024rsprompter}
K.~Chen, C.~Liu, H.~Chen, H.~Zhang, W.~Li, Z.~Zou, and Z.~Shi, ``Rsprompter: Learning to prompt for remote sensing instance segmentation based on visual foundation model,'' \emph{IEEE Transactions on Geoscience and Remote Sensing}, 2024.

\bibitem{ch2017total}
C.~K. Ch'ng and C.~S. Chan, ``Total-{Text}: A comprehensive dataset for scene text detection and recognition,'' in \emph{2017 14th IAPR International Conference on Document Analysis and Recognition (ICDAR)}, vol.~1.\hskip 1em plus 0.5em minus 0.4em\relax IEEE, 2017, pp. 935--942.

\bibitem{singh2021textocr}
A.~Singh, G.~Pang, M.~Toh, J.~Huang, W.~Galuba, and T.~Hassner, ``Text{OCR}: Towards large-scale end-to-end reasoning for arbitrary-shaped scene text,'' in \emph{Proceedings of the IEEE/CVF Conference on Computer Vision and Pattern Recognition}, 2021, pp. 8802--8812.

\bibitem{chng2019icdar2019}
C.~K. Chng, Y.~Liu, Y.~Sun, C.~C. Ng, C.~Luo, Z.~Ni, C.~Fang, S.~Zhang, J.~Han, E.~Ding \emph{et~al.}, ``{ICDAR} 2019 robust reading challenge on arbitrary-shaped text-rrc-art,'' in \emph{2019 International Conference on Document Analysis and Recognition (ICDAR)}.\hskip 1em plus 0.5em minus 0.4em\relax IEEE, 2019, pp. 1571--1576.

\bibitem{craft}
Y.~Baek, B.~Lee, D.~Han, S.~Yun, and H.~Lee, ``Character region awareness for text detection,'' in \emph{Proceedings of the IEEE/CVF conference on computer vision and pattern recognition}, 2019, pp. 9365--9374.

\bibitem{synthtext}
A.~Gupta, A.~Vedaldi, and A.~Zisserman, ``Synthetic data for text localisation in natural images,'' in \emph{IEEE Conference on Computer Vision and Pattern Recognition}, 2016.

\bibitem{dosovitskiy2020image}
A.~Dosovitskiy, L.~Beyer, A.~Kolesnikov, D.~Weissenborn, X.~Zhai, T.~Unterthiner, M.~Dehghani, M.~Minderer, G.~Heigold, S.~Gelly \emph{et~al.}, ``An image is worth 16x16 words: Transformers for image recognition at scale,'' \emph{arXiv preprint arXiv:2010.11929}, 2020.

\bibitem{chen2017deeplab}
L.-C. Chen, G.~Papandreou, I.~Kokkinos, K.~Murphy, and A.~L. Yuille, ``Deeplab: Semantic image segmentation with deep convolutional nets, atrous convolution, and fully connected crfs,'' \emph{IEEE Transactions on Pattern Analysis and Machine Intelligence}, vol.~40, no.~4, pp. 834--848, 2017.

\bibitem{hrnetv2}
J.~Wang, K.~Sun, T.~Cheng, B.~Jiang, C.~Deng, Y.~Zhao, D.~Liu, Y.~Mu, M.~Tan, X.~Wang, W.~Liu, and B.~Xiao, ``Deep high-resolution representation learning for visual recognition,'' \emph{IEEE Transactions on Pattern Analysis and Machine Intelligence}, vol.~43, no.~10, pp. 3349--3364, 2021.

\end{thebibliography}
\end{document}